\documentclass[10pt,twocolumn,letterpaper]{article}
\usepackage[accsupp]{axessibility}

\usepackage{iccv}              

\usepackage{algorithm}
\usepackage{amsmath}
\usepackage{bm}
\usepackage{amssymb}
\usepackage{amsfonts}
\usepackage{cite} %
\usepackage{csquotes} %
\usepackage{physics}
\usepackage{booktabs}
\usepackage{multirow}
\usepackage{xcolor} 
\usepackage{threeparttable}
\usepackage{arydshln}

\usepackage{makecell}
\usepackage{etoolbox}
\usepackage{algpseudocode}
\usepackage{amsmath}

\definecolor{iccvblue}{rgb}{0.21,0.49,0.74}
\usepackage[pagebackref,breaklinks,colorlinks,allcolors=iccvblue]{hyperref}

\title{ViM-VQ: Efficient Post-Training Vector Quantization for Visual Mamba}

\author{
   Juncan Deng$^{1,2}$\thanks{Work done during an internship at vivo Mobile Communication.}, Shuaiting Li$^{1,2}$\footnotemark[1], Zeyu Wang$^{1}$, Kedong Xu$^{2}$,
   Hong Gu$^{2}$, 
   Kejie Huang$^{1}$\thanks{Corresponding Authors}\\ 
   $^{1}$Zhejiang University  \quad \quad
   $^{2}$vivo Mobile Communication Co., Ltd \quad \quad \\
  \small\texttt{\{dengjuncan,list,wangzeyu2020,huangkejie\}@zju.edu.cn}\\
  \small\texttt{\{xukedong, guhong\}@vivo.com} \\
}

\begin{document}
\maketitle
\begin{abstract}
Visual Mamba networks (ViMs) extend the selective state space model (Mamba) to various vision tasks and demonstrate significant potential. As a promising compression technique, vector quantization (VQ) decomposes network weights into codebooks and assignments, significantly reducing memory usage and computational latency, thereby enabling the deployment of ViMs on edge devices. Although existing VQ methods have achieved extremely low-bit quantization (e.g., 3-bit, 2-bit, and 1-bit) in convolutional neural networks and Transformer-based networks, directly applying these methods to ViMs results in unsatisfactory accuracy. We identify several key challenges: 1) The weights of Mamba-based blocks in ViMs contain numerous outliers, significantly amplifying quantization errors. 2) When applied to ViMs, the latest VQ methods suffer from excessive memory consumption, lengthy calibration procedures, and suboptimal performance in the search for optimal codewords. In this paper, we propose ViM-VQ, an efficient post-training vector quantization method tailored for ViMs. ViM-VQ consists of two innovative components: 1) a fast convex combination optimization algorithm that efficiently updates both the convex combinations and the convex hulls to search for optimal codewords, and 2) an incremental vector quantization strategy that incrementally confirms optimal codewords to mitigate truncation errors. Experimental results demonstrate that ViM-VQ achieves state-of-the-art performance in low-bit quantization across various visual tasks.
\end{abstract}

\section{Introduction}
\label{sec:intro}

Recent advancements have sparked growing research interest in state space models (SSMs)~\citep{kalman1960new}. Modern SSMs are particularly adept at capturing long-range dependencies and benefit from parallel training capabilities. Several SSM-based methods, including the structured state space sequence model (S4)~\citep{gu2021efficiently} and S4D~\citep{gu2022parameterization}, have been introduced to efficiently handle long sequence data across various tasks, owing to their convolutional computations and near-linear complexity. The recent development of Mamba~\citep{gu2023mamba} further extends the SSM framework by incorporating time-varying parameters and proposing a hardware-aware algorithm for highly efficient training and inference. This has sparked increasing interest in adapting Mamba to vision tasks~\citep{zhu2024vision, liu2024vMamba, huang2024localMamba, yang2024plainmamba}, giving rise to the concept of Visual Mamba networks (ViMs). Among these ViMs, Vision Mamba (Vim) and VMamba have gained particular attention due to their effectiveness and unique architectures. 

Model quantization compresses deep neural networks (DNNs), enabling their deployment and inference on resource-constrained devices. Quantization methods are generally classified into uniform quantization (UQ)~\citep{krishnamoorthi2018quantizing, esser2019learned, nagel2020up, li2021brecq, xiao2023smoothquant, frantar2022gptq, lin2024awq, shao2023omniquant} and vector quantization (VQ) ~\citep{stock2019and, martinez2021permute, cho2021dkm, yvinec2023network, van2024gptvq, deng2024vq4dit, egiazarian2024extreme, tseng2024quip}. UQ reduces the precision of network weights, while VQ clusters network weights into codebooks and corresponding assignments. Among various quantization paradigms, post-training quantization (PTQ)~\citep{liu2021post, shang2023post} is a particularly popular technique, as it rapidly calibrates quantization parameters using a small dataset without the need for full retraining or time-consuming fine-tuning. Both UQ and VQ have been widely applied to convolutional neural networks~\citep{o2015introduction} and Transformer-based networks~\citep{waswani2017attention}, with research on UQ in ViMs already underway~\citep{xu2025mambaquant, cho2024ptq4vm}. Nevertheless, the feasibility of achieving extreme compression via VQ for ViMs remains unexplored.

\begin{figure*}[t]
  \centering
  \includegraphics[width=1\linewidth]{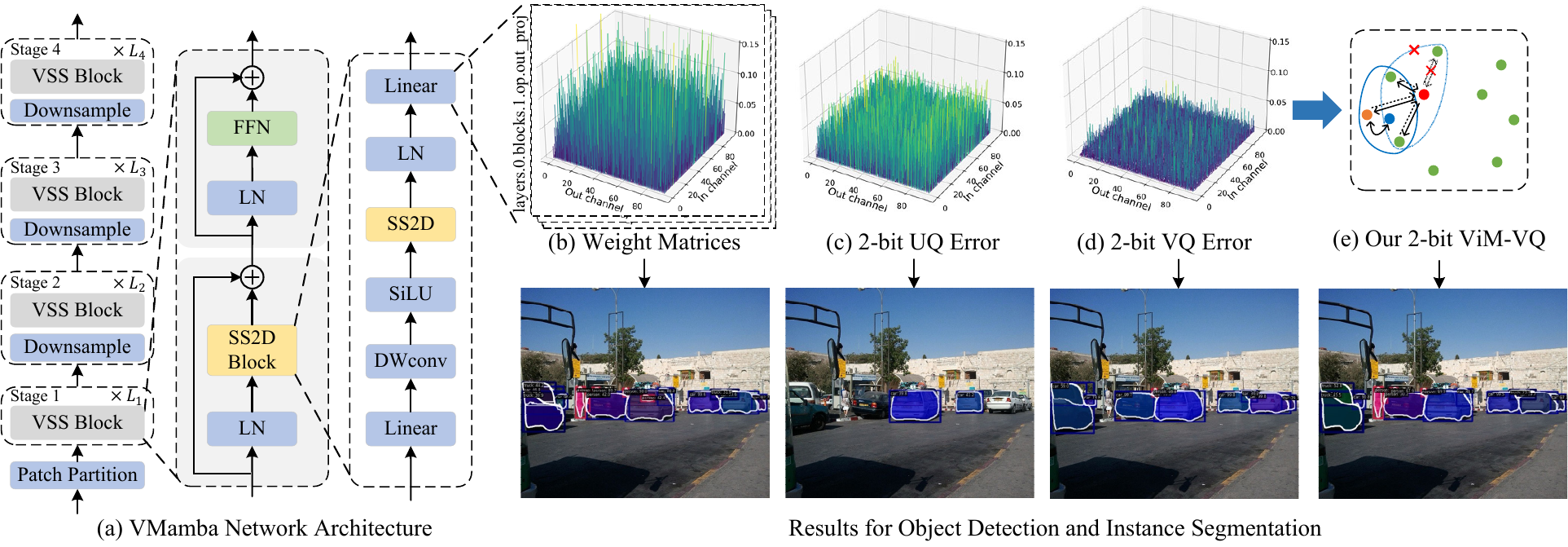}
    \caption{Visualization of the weight distribution and quantization error for the Mamba-based blocks within Visual Mamba networks (ViMs). (a) VMamba~\citep{liu2024vMamba} comprises numerous Mamba-based~\citep{gu2023mamba} blocks (e.g., VSS blocks). (b) The weights of these blocks exhibit significant outliers. (c-d) Although conventional vector quantization (VQ) outperforms uniform quantization (UQ) at low-bit precision, substantial quantization errors still severely degrade network performance. (e) Our ViM-VQ restores performance to a level nearly identical to the full-precision model. }
  \label{error}
  \captionsetup{skip=2pt}
  \vspace{-0.2cm}
\end{figure*}

We first identify several critical challenges encountered when applying VQ to ViMs: 
\textbf{1) Outliers in ViMs' Weights.} As shown in Figure~\ref{error}, the weights of the Mamba-based~\citep{gu2023mamba} blocks in ViMs exhibit significant outliers. The presence of outliers leads to considerable quantization errors, resulting in unsatisfactory accuracy. 
\textbf{2) Limitations of Existing VQ Methods.} As depicted in Figure~\ref{pipeline}, existing VQ methods for visual networks suffer from substantial calibration steps, significant GPU memory consumption, and suboptimal performance when applied to multi-layered networks such as ViMs. 
\textbf{3) Truncation Error from Assignment Conversion.} Figure~\ref{ratio} illustrates that weighted average methods suffer from truncation errors introduced during the soft-to-hard assignment conversion, causing degradation of inference performance relative to calibration.

To address these challenges in quantizing ViMs, we propose ViM-VQ, an efficient post-training vector quantization method for Visual Mamba networks (ViMs). ViM-VQ introduces a fast convex combination optimization algorithm that efficiently initializes and updates both the convex combinations and hulls, rapidly searching for optimal codewords while significantly reducing computational and memory overhead. Additionally, ViM-VQ adopts an incremental vector quantization strategy that progressively confirms optimal codewords, thereby mitigating truncation errors. Our major contributions are summarized as follows:

\begin{itemize}
    \item ViM-VQ efficiently quantizes ViMs, achieving superior performance while significantly reducing both calibration time and GPU memory requirements.
    \item To the best of our knowledge, our ViM-VQ is the first vector quantization framework specifically designed for Visual Mamba networks (ViMs). We analyze ViMs' weight distribution and address the unique challenges of applying VQ to ViMs through targeted improvements. 
    \item Experiments demonstrate that, compared to other advanced VQ methods, our ViM-VQ achieves superior performance in low-bit quantization across various tasks.
\end{itemize}

\section{Related Work}
\label{sec:related}

\paragraph{Visual Mamba Networks.} Mamba~\citep{gu2023mamba} is an advanced structured state space model that significantly enhances the capabilities of state space models (SSMs) for processing sequential data. It transforms the parameters of the structured state space model (S4)~\citep{gu2021efficiently} into learnable functions, introducing a parallel scanning method to improve efficiency. By overcoming the local perception limitations of convolutional neural networks (CNNs)~\citep{o2015introduction} and the quadratic computational complexity of Transformers~\citep{waswani2017attention}, Visual Mamba networks (ViMs)~\citep{liu2024vision} have found widespread success across a range of visual tasks. Vision Mamba (Vim)~\citep{zhu2024vision} represents the first use of Mamba in computer vision, utilizing bidirectional SSMs for global feature modeling and position embeddings to enhance spatial awareness. Subsequently, VMamba~\citep{liu2024vMamba} introduces a cross-scan module to address direction-sensitive challenges. LocalMamba~\citep{huang2024localMamba} improves performance by incorporating local inductive biases, while PlainMamba~\citep{yang2024plainmamba} adopts a non-hierarchical structure for better multi-scale integration. Mamba-ND~\citep{li2024mamba} extends the Mamba architecture to multi-dimensional data by altering the sequence order. Despite the impressive performance of ViMs in computer vision tasks, their large size remains a significant obstacle for deployment on edge devices with limited resources.

\paragraph{Model Quantization.} Model quantization is an effective technique to compress deep neural networks by reducing their model size and improving inference efficiency. Current quantization methods can be broadly classified into two main categories~\citep{van2024gptvq}: uniform quantization (UQ), which converts model weight from high to low precision, and vector quantization (VQ), which decomposes weight into a codebook and corresponding assignments. Most quantization approaches are implemented via post-training quantization (PTQ)~\citep{liu2021post, shang2023post}, which requires only a small calibration dataset.
In the UQ domain, GPTQ~\citep{frantar2022gptq} proposes a layer-wise quantization method that leverages approximate second-order information to quantize weights with minimal accuracy loss. Recent studies have begun exploring UQ methods specifically tailored for ViMs. PTQ4VM~\citep{cho2024ptq4vm} introduces per-token static quantization and joint learning of smoothing scale and step size to mitigate outliers. MambaQuant~\citep{xu2025mambaquant} applies the Hadamard transformation to standardize the variance across channels.
In the VQ domain, PQF~\citep{martinez2021permute} utilizes rate-distortion theory to reorder weights, thereby reducing clustering errors. DKM~\citep{cho2021dkm} integrates a differentiable K-Means formulation into the objective function to preserve the network's accuracy. JLCM~\citep{yvinec2023network} builds upon DKM by adjusting the gradient computation for assignments to avoid incorrect updates. VQ4DiT~\citep{deng2024vq4dit} identifies that fixed initial assignments can lead to suboptimal updates, proposing an improved search algorithm to dynamically select more appropriate codewords.
Although these methods have shown promising results in CNNs and Transformers, their performance under extremely low-bit quantization settings remains limited in ViMs. To the best of our knowledge, our approach is the first post-training vector quantization solution explicitly designed for ViMs.

\section{Preliminaries}
\label{sec:prelim}

\subsection{Mamba}
Mamba~\citep{gu2023mamba} is a state space model (SSM) inspired by continuous systems, mapping a 1-D sequence $x(t) \in \mathbb{R} $ to $ y(t) \in \mathbb{R}$ through a hidden state $h(t) \in \mathbb{R}^\mathtt{N}$. This system uses $\mathbf{E} \in \mathbb{R}^{\mathtt{N} \times \mathtt{N}}$ as the evolution matrix and $\mathbf{B} \in \mathbb{R}^{\mathtt{N} \times 1}$, $\mathbf{P} \in \mathbb{R}^{1 \times \mathtt{N}}$ as the projection matrices. The continuous systems work as follows: $h'(t) = \mathbf{E}h(t) + \mathbf{B}x(t)$ and $y(t) = \mathbf{P}h(t)$.
Mamba introduces a timescale parameter $\mathbf{\Delta}$ to transform $\mathbf{E}$ and $\mathbf{B}$ to discrete matrices $\mathbf{\overline{E}}$ and $\mathbf{\overline{B}}$. This transformation employs the zero-order hold method, defined as:
\begin{equation}
\begin{aligned}
\mathbf{\overline{E}} &= \exp(\mathbf{\Delta}\mathbf{E}), \\
\mathbf{\overline{B}} &= (\mathbf{\Delta} \mathbf{E})^{-1}(\exp(\mathbf{\Delta} \mathbf{E}) - \mathbf{I}) \cdot \mathbf{\Delta} \mathbf{B}.
\end{aligned}
\end{equation}
With these discrete matrices, the system dynamics and the output computed via a global convolution can be respectively rewritten as:
\begin{equation}
\begin{aligned}
y_t &= \mathbf{P}h_t = \mathbf{P}(\mathbf{\overline{E}}h_{t-1} + \mathbf{\overline{B}}x_{t}), \\
\mathbf{y} &= \mathbf{x} * \mathbf{\overline{K}} = \mathbf{x} * (\mathbf{P}\mathbf{\overline{B}}, \mathbf{P}\mathbf{\overline{E}}\mathbf{\overline{B}}, \dots, \mathbf{P}\mathbf{\overline{E}}^{\mathtt{M}-1}\mathbf{\overline{B}}),
\end{aligned}
\end{equation}
where $\mathtt{M}$ is the length of the input sequence $\mathbf{x}$, and $\overline{\mathbf{K}} \in \mathbb{R}^{\mathtt{M}}$ is a structured convolutional kernel.

\subsection{Visual Mamba Networks}
Vision Mamba (Vim)~\citep{zhu2024vision} and VMamba~\citep{liu2024vMamba} are the most commonly used Visual Mamba networks (ViMs)~\citep{liu2024vision}. Inspired by ViT~\citep{dosovitskiy2020image} and BERT~\citep{devlin2018bert}, Vim is the first network to extend Mamba to vision tasks. Vim incorporates position embeddings for annotating image sequences and utilizes bidirectional state space models to efficiently compress visual representations. VMamba achieves superior performance compared to Vim. It processes input images by first partitioning them into patches through a stem module, yielding an initial 2D feature map. Without additional positional embeddings, VMamba progressively constructs hierarchical features across multiple stages, each comprising a downsampling layer and several Visual State Space (VSS) blocks analogous to Mamba-based blocks.

\subsection{Vector Quantization}
Vector quantization (VQ) offers more flexibility than uniform quantization (UQ), making it especially suitable for extremely low-bit quantization scenarios. VQ typically employs clustering algorithms such as K-Means~\citep{hartigan1979algorithm} to decompose the weight matrix $ \mathbf{W} \in \mathbb{R}^{o \times i} $ into a codebook and assignments.  
Specifically, $ \mathbf{W} $ is divided into $ d $-dimensional sub-vectors $ \mathbf{W} = \{ w_{o,i/d} \} $, where $ o \times i / d $ is the total number of sub-vectors.  
These sub-vectors are then represented using a codebook $ \mathbf{C} \in \mathbb{R}^{k \times d} $, where $ c(k) $ denotes the $ k $-th codeword of length $ d $. The assignments $ \mathbf{A} = \{ a_{o,i/d} \} $ indicate the index of the codeword that best represents each sub-vector.  
Finally, the quantized weight $ \widehat{\mathbf{W}} $ is reconstructed by replacing each $ w_{o,i/d} $ with $ c(a_{o,i/d}) $:
\begin{equation}
\widehat{\mathbf{W}} = \mathbf{C}[\mathbf{A}] =
\begin{bmatrix}
    c(a_{1,1}) & c(a_{1,2}) & \cdots & c(a_{1,i/d}) \\
    c(a_{2,1}) & c(a_{2,2}) & \cdots & c(a_{2,i/d}) \\
    \vdots & \vdots & \ddots & \vdots \\
    c(a_{o,1}) & c(a_{o,2}) & \cdots & c(a_{o,i/d})
\end{bmatrix}.
\label{cfunction}
\end{equation}
Assignments $ \mathbf{A} $ can be stored using $ \frac{o \times i}{d} \times \log_2 k $ bits, and codebook $ \mathbf{C} $ can be stored using $ k \times d \times 32 $ bits (typically negligible). Therefore, the quantization bit for quantizing a 32-bit original weight can be computed as $\frac{\log_2 k}{d}$.

\begin{figure*}[t]
  \centering
  \includegraphics[width=1.0\linewidth]{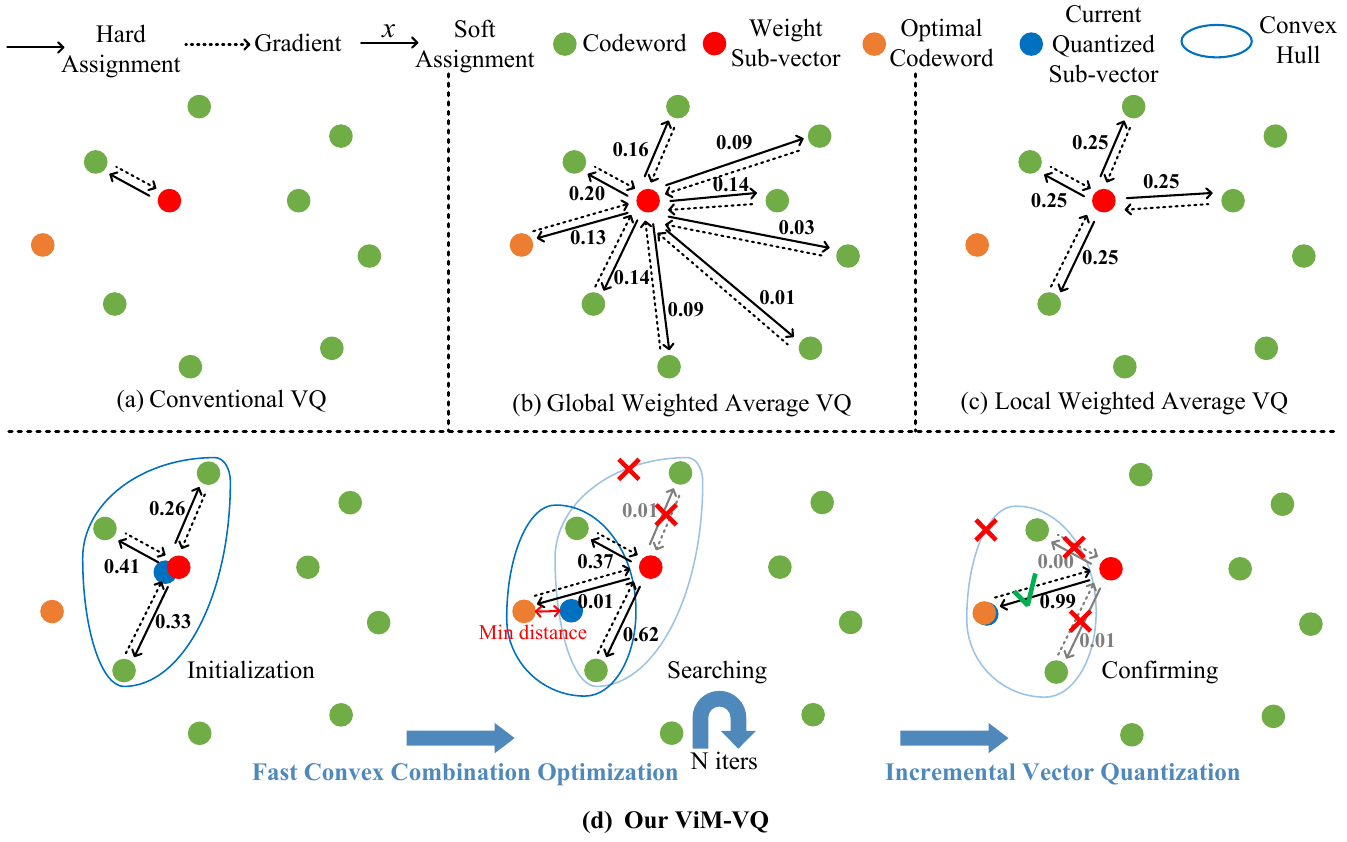}
    \caption{Comparison of different vector quantization (VQ) algorithms. (a) Conventional VQ algorithms~\citep{deepcomp_iclr16, stock2019and, martinez2021permute} rely on simple distance-based assignment, making them unsuitable for the task. (b) Global weighted average VQ algorithms~\citep{cho2021dkm, yvinec2023network} use and update soft assignments to all codewords and select the maximum as the final assignment. However, they suffer from high calibration and memory overhead, especially for multi-layered structures such as Visual Mamba networks (ViMs)~\citep{zhu2024vision, liu2024vMamba}. (c) Local weighted average VQ algorithms~\citep{deng2024vq4dit} reduce overhead by limiting soft assignments to a few nearest codewords but compromise performance. (d) Our ViM-VQ overcomes these limitations by using a fast convex combination optimization algorithm and an incremental vector quantization strategy to efficiently identify optimal codewords while compensating for truncation errors.}
  \label{pipeline}
  \captionsetup{skip=2pt}
  \vspace{-0.2cm}
\end{figure*}

\section{Method}
\label{sec:method}

\subsection{Motivation}

Quantizing the weights of Visual Mamba networks (ViMs) to extremely low-bit precision significantly reduces memory usage and accelerates inference, making ViMs suitable for deployment on resource-constrained edge devices. However, as illustrated in Figure~\ref{error}, the weights of ViMs contain numerous outliers, leading to \textbf{substantial quantization errors} when applying conventional uniform quantization (UQ) or vector quantization (VQ), ultimately resulting in model performance degradation. The observation that VQ incurs a relatively smaller quantization error underscores its superiority over UQ in extremely low-bit scenarios~\citep{van2024gptvq, deng2024vq4dit}, thus motivating our focus on applying VQ to ViMs. As illustrated in Figure~\ref{pipeline}, existing VQ approaches exhibit the following limitations when applied to ViMs:

\textbf{Conventional VQ.} Deep compression~\citep{deepcomp_iclr16}, ATB~\citep{stock2019and}, and PQF~\citep{martinez2021permute} fail to adequately adjust the distribution of quantized weights to support extremely low-bit quantization. Consequently, these methods have primarily been effective for smaller networks (e.g., ResNet-18~\citep{he2016deep}) and have shown limited scalability for larger models like ViMs.

\textbf{Global Weighted Average VQ.} DKM~\citep{cho2021dkm} and JLCM~\citep{yvinec2023network} demand not only substantial GPU memory for storing and updating soft assignment ratios but also significant calibration steps. These limitations render them impractical for multi-layered architectures such as ViMs. For instance, as indicated in Table~\ref{metric}, using a $256 \times 4$ codebook for 2-bit quantization significantly increases the number of learnable parameters by $256 \times o \times i / 4$ per layer, potentially causing out-of-memory issues on resource-constrained devices.

\begin{figure}[t]
  \centering
  \includegraphics[width=1.0\linewidth]{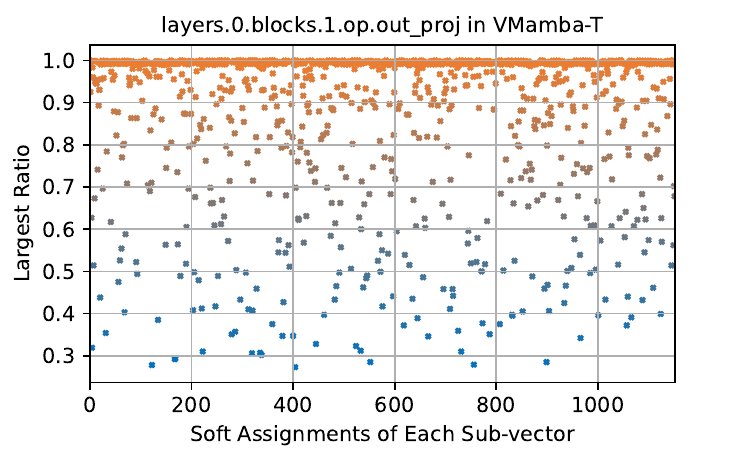}
    \caption{Distribution of maximum soft assignment ratios in weighted average vector quantization (4 candidates per sub-vector, $256 \times 8$ codebook). When these maximum ratios deviate significantly from 1.0, it indicates two issues: (1) optimal codewords lie outside current candidates, limiting performance; (2) hard assignment conversion causes significant truncation errors.}
  \label{ratio}
  \captionsetup{skip=2pt}
  \vspace{-0.3cm}
\end{figure}

\textbf{Local Weighted Average VQ.} Although VQ4DiT~\citep{deng2024vq4dit} addresses the memory and computational overhead issues present in earlier methods, its performance remains suboptimal due to the limited scope of candidate codewords (see Figure~\ref{ratio}). Moreover, their approach of initializing soft assignments with uniform values leads to a substantial initial discrepancy between the quantized and original weights, which can cause gradient oscillations during optimization.

\textbf{Truncation Error from Assignment Conversion.} As demonstrated in Figure~\ref{ratio} and Table~\ref{metric}, existing weighted average VQ methods neglect the truncation error that is introduced when converting from soft to hard assignments to finalize the optimal codewords. This oversight leads to noticeable inference accuracy degradation compared to calibration results.

Therefore, we propose ViM-VQ, a fast post-training vector quantization method specifically tailored for the extremely low-bit quantization of ViMs. Our method consists of two innovative components:
1) To overcome the challenges of high GPU memory usage, extensive calibration steps, and suboptimal performance encountered by weighted average VQ methods, we develop a \textbf{fast convex combination optimization} algorithm. This algorithm operates on convex combinations formed by several nearby candidate codewords and their corresponding soft assignment ratios. Both the convex combinations and the convex hulls are rapidly optimized, significantly accelerating the search for optimal codewords.
2) To eliminate the truncation error resulting from assignment conversion, we develop an \textbf{incremental vector quantization} strategy, in which soft assignments with high ratios are transitioned from soft to hard to confirm optimal codewords in each calibration iteration. This strategy allows the sub-vectors still undergoing calibration to compensate for truncation errors, thereby preventing the accumulation of errors throughout the network.

\subsection{Fast Convex Combination Optimization}
The key challenge in VQ is determining the optimal codewords to accurately represent each weight sub-vector. As illustrated in Figure~\ref{pipeline}(d), given a codebook $\mathbf{C}$ generated via the K-Means algorithm~\citep{hartigan1979algorithm}, we propose a fast convex combination optimization algorithm that begins by identifying $\text{top-}n$ nearest candidate codewords based on the Euclidean distance from each sub-vector:
\begin{equation}
  \mathcal{C}_{o,i/d} \kern-2pt=\kern-2pt \left\{ c_{o,i/d}^1, \dots, c_{o,i/d}^n \right\} \kern-2pt=\kern-2pt \operatorname{Top}_n (-\| w_{o,i/d} - c(k) \|_2^2),
\end{equation}
where $\mathcal{C}_{o,i/d}$ is the set of candidate codewords for sub-vector $w_{o,i/d}$. Next,  
we compute the soft assignment ratios $\mathcal{R}_{o,i/d} \kern-2pt=\kern-2pt \left\{ r_{o,i/d}^1, \dots, r_{o,i/d}^n \right\}$ for these candidates using the softmax function on a set of learnable scores $\{z_{o,i/d}^m\}_{m=1}^n:$
\begin{equation} 
r_{o,i/d}^m = \frac{\exp\left({z_{o,i/d}^m}\right)}{\sum_{j=1}^n \exp\left({z_{o,i/d}^j}\right)}, \quad \text{where} \sum_{m=1}^n r_{o,i/d}^m = 1. 
\end{equation}
Here, $z_{o,i/d}^m$ represents the unnormalized score for the $m$-th ratio.
In this way, the learnable $\mathcal{C}_{o,i/d}$ along with the learnable $\mathcal{R}_{o,i/d}$ form differentiable convex combinations that produce the quantized weight $\widehat{\mathbf{W}}$ as follows:
\begin{equation}
    \widehat{\mathbf{W}} \kern-5pt=\kern-5pt 
    \begin{bmatrix}
        \widehat{w}_{1,1}\!=\!\sum \mathcal{C}_{1,1} \mathcal{R}_{1,1} & \!\cdots\! & \widehat{w}_{1,i/d}\!=\!\sum \mathcal{C}_{1,i/d} \mathcal{R}_{1,i/d}\\
        \widehat{w}_{2,1}\!=\!\sum \mathcal{C}_{2,1} \mathcal{R}_{2,1} & \!\cdots\! & \widehat{w}_{2,i/d}\!=\!\sum \mathcal{C}_{2,i/d} \mathcal{R}_{2,i/d}\\
        \vdots  & \ddots & \vdots \\
        \widehat{w}_{o,1}\!=\!\sum \mathcal{C}_{o,1} \mathcal{R}_{o,1} & \!\cdots\! & \widehat{w}_{o,i/d}\!=\!\sum \mathcal{C}_{o,i/d} \mathcal{R}_{o,i/d}\\
    \end{bmatrix}\kern-3pt.
\end{equation}

The combinations can be updated via gradient descent:
\begin{equation}
\begin{aligned}
\mathcal{C}_{o,i/d} &\leftarrow \mathcal{C}_{o,i/d} - O(\nabla_{\mathcal{C}_{o,i/d}} \mathcal{L_\text{init}}, \theta_c), \\
\mathcal{R}_{o,i/d} &\leftarrow \mathcal{R}_{o,i/d} - O(\nabla_{\mathcal{R}_{o,i/d}} \mathcal{L_\text{init}}, \theta_r), 
\end{aligned}
\end{equation}
where $O(,)$ is the optimizer with hyperparameter $\theta$.
For the initialization of them, we optimize their corresponding codewords $c_{o,i/d}^m$ and ratios $r_{o,i/d}^m$ to minimize the reconstruction error $\mathcal{L_\text{init}} = \| \mathbf{W} - \widehat{\mathbf{W}} \|_2^2$.
This initialization ensures that $ \widehat{\mathbf{W}} $ is initially close to $ \mathbf{W} $, thus preventing severe gradient oscillations during subsequent calibration.
 
The convex combination optimization facilitates the search for optimal codewords. To ensure optimal codewords are contained within effective convex hulls, we introduce an adaptive replacement strategy. The key insight is that if a current quantized sub-vector $\widehat{w}_{o,i/d}$ is distant from certain candidates (i.e., those with near-zero ratios), the optimal codeword is likely not near them.
Consequently, we replace these irrelevant candidates with new candidate codewords that are closest to $\widehat{w}_{o,i/d}$:
\begin{equation}
    c_{o,i/d}^m \leftarrow \underset{c(k) \in \mathbf{C} \setminus \mathcal{C}_{o,i/d}}{\operatorname{argmin}} \| \widehat{w}_{o,i/d} - c(k) \|_2^2,
\end{equation}
where $r_{o,i/d}^m\approx0$ is smaller than $\lambda$ (e.g., 1e-2).
This adaptive replacement procedure dynamically shifts the convex hulls towards more optimal codewords without introducing significant perturbations to the network's outputs.

The key differences between the fast convex combination optimization algorithm and previous works can be summarized as follows:

\begin{itemize}
    \item Unlike global methods~\citep{cho2021dkm, yvinec2023network}, which incur high calibration costs and memory overhead, or local methods~\citep{deng2024vq4dit}, which yield suboptimal quantization, our approach strikes an effective balance. The simultaneous optimization of convex combinations (ratios) and convex hulls (candidate codewords) accelerates the search process and reduces GPU memory usage, resulting in superior performance.

    \item Rather than using hard assignments and approximated gradients~\citep{deepcomp_iclr16, haq, lee2021network, stock2019and, martinez2021permute}, our algorithm leverages differentiable convex combinations directly based on the objective function of the task.
    
    \item Our algorithm is a fully differentiable process and does not require expensive secondary computations such as Hessian trace~\citep{hawq-v2, limit_quant, liu2024vptq, egiazarian2024extreme} or SVD~\citep{lee2021network,pmlr-v80-wu18h, zhu2023learning}.
\end{itemize}

\begin{figure}[t]
  \centering
  \includegraphics[width=0.95\linewidth]{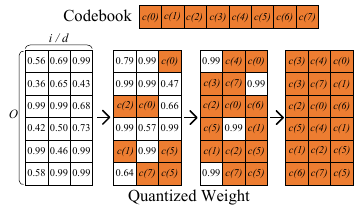}
    \caption{Incremental vector quantization strategy of ViM-VQ. High-ratio soft assignments are converted to hard assignments to confirm optimal codewords, while remaining soft assignments compensate for truncation errors during calibration.}
  \label{incremental}
  \captionsetup{skip=2pt}
  \vspace{-0.2cm}
\end{figure}

\subsection{Incremental Vector Quantization}
As shown in Figure~\ref{ratio}, weighted average VQ algorithms enforce a one-time soft-to-hard transition in all assignments to confirm the optimal codewords, which introduces truncation errors into quantized weights:
\begin{equation}
    \left\| c_{o,i/d}^{m^*} - \sum \mathcal{C}_{o,i/d} \mathcal{R}_{o,i/d} \right\|_2 > 0, m^* = \underset{m}{\operatorname{argmax}} \, r_{o,i/d}^m.
\end{equation}
Table~\ref{metric} further demonstrates that these truncation errors cause deviations in intermediate features within each block of ViMs, ultimately accumulating and leading to decreased accuracy in the final output.

Inspired by INQ~\citep{zhou2017incremental}, we propose an incremental vector quantization strategy. As illustrated in Figure~\ref{incremental}, our strategy confirms candidate codewords with soft assignment ratios close to 1.0 (e.g., $>$ 0.99) as optimal codewords at each calibration step:
\begin{equation} 
\widehat{w}_{o,i/d} \leftarrow c_{o,i/d}^m, \quad a_{o,i/d} \leftarrow \text{index}(c_{o,i/d}^m). 
\end{equation}

Compared to previous methods that perform a one-time assignment conversion, our incremental strategy significantly reduces truncation errors at each calibration step. This allows the remaining weights still undergoing calibration to compensate for these quantization errors, thereby effectively recovering the network's accuracy. Moreover, our method ensures consistent network performance between the calibration and inference phases.

The strategy employs a combination of three objective functions to efficiently identify and confirm optimal codewords. Specifically, given the input $\mathbf{x}$ and the target $\mathbf{y}$ from the calibration dataset for the quantized ViM model $\mathbf{\epsilon}_q$, the task objective function is defined as:
\begin{equation}
    \mathcal{L}_{t} = \mathbb{E}_{\mathbf{x},\mathbf{y}}\left[ \| \mathbf{y} - \mathbf{\epsilon}_q(\mathbf{x}) \|_2^2 \right].
\end{equation}

To ensure consistency in the Mamba-based block outputs between the original and quantized networks, we employ a block-wise knowledge distillation (BKD) objective:
\begin{equation}
    \mathcal{L}_{bkd} = \mathbb{E}_{\mathbf{x}}\left[ \sum_{l} \| b_{fp}^{l}(\mathbf{x}) - b_q^{l}(\mathbf{x}) \|_2^2 \right],
\end{equation}
where $b_{fp}^l$ and $b_q^l$ represent the output features of the $l$-th block from the full-precision model $\mathbf{\epsilon}_{fp}$ and the quantized model $\mathbf{\epsilon}_q$, respectively.

Additionally, we introduce a regularization term $\mathcal{L}_r$ to encourage rapid convergence of convex combinations:
\begin{equation} 
\mathcal{L}_r =  \left( \frac{d}{o \times i} \right) \sum_{o,i/d} \sum_{m=1}^n r_{o,i/d}^m \left(1 - r_{o,i/d}^m \right). 
\end{equation}

The final objective function $\mathcal{L}$ is formulated as:
\begin{equation}
    \mathcal{L} = \mathcal{L}_t + \mathcal{L}_{bkd} + 
    \begin{cases}
    \mathcal{L}_r, & \text{if } \mathcal{L}_r^{(t)} > \mathcal{L}_r^{(t-1)},\\[0.5em]
    0, & \text{otherwise},
    \end{cases}
\end{equation}
where $\mathcal{L}_r$ is selectively included only when it increases compared to the previous iteration. This selective incorporation ensures stable updates and convergence of the differentiable convex combinations.

\section{Experiment}
\label{sec:experiment}

Our experiments are conducted on two Visual Mamba networks (ViMs)~\citep{liu2024vision}: Vision Mamba (Vim)~\citep{zhu2024vision} and VMamba~\citep{liu2024vMamba}. The experiments cover four main computer vision tasks: image classification, semantic segmentation, object detection, and instance segmentation. Our primary goal is to investigate the trade-off between model weight quantization and accuracy. Therefore, activation quantization is disabled in all experiments. We sample 100 images per category from the training set to form the calibration set, using a batch size of 128. All experiments are conducted on a single NVIDIA A6000 GPU with 48GB of memory.

\textbf{Hyperparameters:} The codebook sizes for each layer in the 3/2/1-bit quantization are set to $2^{6} \times 2$, $2^{8} \times 4$, and $2^{8} \times 8$, respectively. The codebooks and ratios of soft assignments are updated using the Adamax optimizer with a fixed learning rate of $1 \times 10^{-5}$ and $5 \times 10^{-2}$, respectively, without additional hyperparameter tuning. Dataset processing and other hyperparameters follow the original settings of ViMs. Consistent with all baselines, we quantize only the linear projection layers within the ViMs blocks, as they contribute significantly to memory usage.

\textbf{Baselines:} We compare our method against a broad range of uniform quantization (UQ) and vector quantization (VQ) methods. UQ methods include basic uniform quantization (Round-to-Nearest), GPTQ~\citep{frantar2022gptq}, and recent ViMs-specific quantization techniques such as MambaQuant~\citep{xu2025mambaquant} and PTQ4VM~\citep{cho2024ptq4vm}. The recent VQ methods designed for visual networks include the basic vector quantization (K-Means~\citep{hartigan1979algorithm}), PQF~\citep{martinez2021permute}, DKM~\citep{cho2021dkm}, and VQ4DiT~\citep{deng2024vq4dit}.

\begin{table}[t]
    \centering
    \small
    \setlength{\tabcolsep}{5pt}
    \begin{tabular}{c|c|c|ccc}
    \toprule
    \multirow{2}{*}{\begin{tabular}[c]{@{}c@{}}Model \\ (Params)\end{tabular}} & \multirow{2}{*}{Method} & \multicolumn{4}{c}{Top-1 Accuracy (\%)} \\
                              &  & FP & 3-bit & 2-bit & 1-bit \\
    \midrule
    \multirow{9}{*}{\begin{tabular}[c]{@{}c@{}}Vim-T \\ (7M)\end{tabular}}
          & $\text{UQ} ^ \dagger $ & \multirow{9}{*}{76.07} & 52.84 & 0.16 & - \\
          & GPTQ &  & 56.74 & 0.17 & - \\
          & MambaQuant &  & 57.21 & 0.18 & - \\
          & PTQ4VM &  & 57.29 & 0.18 & - \\
          & $\text{VQ} ^ \dagger $ &  & 67.56 & 33.18 & 0.22 \\
          & PQF &  & 71.71 & 61.84 & 4.20 \\
          & DKM &  & 73.64 & 71.25 & 68.29 \\
          & VQ4DiT &  & 73.57 & 71.04 & 67.72 \\
          & ViM-VQ (Ours) &  & \textbf{74.79} & \textbf{72.17} & \textbf{69.93} \\
    \midrule
    \multirow{9}{*}{\begin{tabular}[c]{@{}c@{}}Vim-S \\ (26M)\end{tabular}}
          & $\text{UQ} ^ \dagger $ & \multirow{9}{*}{80.48} & 72.31 & 0.16 & - \\
          & GPTQ &  & 74.45 & 0.20 & - \\
          & MambaQuant &  & 74.57 & 0.24 & - \\
          & PTQ4VM &  & 74.68 & 0.25 & - \\
          & $\text{VQ} ^ \dagger $ &  & 77.59 & 65.96 & 0.37 \\
          & PQF &  & 78.46 & 73.28 & 10.93 \\
          & DKM &  & 79.70 & 77.96 & 70.38 \\
          & VQ4DiT &  & 79.37 & 77.87 & 69.73 \\
          & ViM-VQ (Ours) &  & \textbf{80.10} & \textbf{78.66} & \textbf{72.02} \\
    \midrule
    \multirow{9}{*}{\begin{tabular}[c]{@{}c@{}}Vim-B \\ (98M)\end{tabular}}
          & $\text{UQ} ^ \dagger $ & \multirow{9}{*}{81.88} & 76.17 &2.06 & - \\
          & GPTQ &  & 76.81 & 2.47 & - \\
          & MambaQuant &  & 77.01 & 3.49 & - \\
          & PTQ4VM &  & 77.17 & 3.62 & - \\
          & $\text{VQ} ^ \dagger $ &  & 77.78 & 70.88 & 5.28 \\
          & PQF &  & 78.76 & 76.02 & 27.40 \\
          & DKM &  & 80.02 & 79.18 & 74.57 \\
          & VQ4DiT &  & 79.73 & 78.86 & 74.37 \\
          & ViM-VQ (Ours) &  & \textbf{80.34} & \textbf{79.46} & \textbf{75.58} \\
    \midrule
    \end{tabular}
    \caption{Quantization results of Vision Mamba (Vim) for image classification on ImageNet-1K. $\text{UQ} ^ \dagger $ and $\text{VQ} ^ \dagger $ represent basic uniform quantization and vector quantization, respectively.}
    \label{cla_result}
    \captionsetup{skip=4pt}
    \vspace{-0.2cm}
\end{table}

\subsection{Image Classification}
Table~\ref{cla_result} presents the image classification performance on three Vision Mamba models of different scales (Vim-T/S/B) evaluated on ImageNet-1K. Uniform quantization (UQ) methods collapse at 2-bit quantization due to significant quantization errors, even with the latest Vim-specific approaches, such as MambaQuant and PTQ4VM. Across all model scales and bit-widths, our ViM-VQ consistently achieves state-of-the-art quantization performance, maintaining the highest Top-1 accuracy. Specifically, for ViM-T, ViM-VQ achieves accuracies of 74.79\% (3-bit), 72.17\% (2-bit), and 69.93\% (1-bit), clearly outperforming other advanced VQ methods like DKM and VQ4DiT. These trends hold for the larger ViM-S and ViM-B models, where ViM-VQ's performance remains remarkably close to the full-precision baseline, demonstrating its robustness even under extremely low-bit conditions.

\begin{table}[t]
\centering
\small
\renewcommand{\arraystretch}{0.9}
\begin{tabular}{l|c|cccc}
\toprule
Bit & Method & {\begin{tabular}[c]{@{}c@{}}mIoU \\ (SS)\end{tabular}} & {\begin{tabular}[c]{@{}c@{}}mIoU \\ (MS)\end{tabular}} & mAcc & aAcc  \\ 
\midrule
 \multirow{3}{*}{FP} & Swin-T & 44.5 & 45.8 & - & -  \\ 
  & ConvNeXt-T & 46.0 & 46.7  & - & - \\ 
  & VMamba-T & 47.9 & 48.8 & 59.2 & 82.4  \\ 
\midrule
\multirow{4}{*}{2}& PQF    & 42.7 & 43.6 & 58.2 & 79.7 \\ 
& DKM   & 44.1 & 44.8 & 58.7 & 80.6  \\  
& VQ4DiT    & 43.9 & 44.7 &  58.4 & 80.6 \\ 
& ViM-VQ (Ours)  & \textbf{44.4} & \textbf{45.4} & \textbf{58.9} & \textbf{81.0} \\
\midrule
\multirow{4}{*}{1}& PQF   & 28.5 & 29.0 & 39.9 & 72.6  \\ 
& DKM   & 37.2 & 37.6 & 52.8 & 76.6  \\  
& VQ4DiT   & 36.2 & 36.8 & 51.9 & 76.2  \\ 
& ViM-VQ (Ours)  & \textbf{41.1} & \textbf{41.8} & \textbf{56.0} & \textbf{77.3} \\ 
\bottomrule
\end{tabular}
\caption{Quantization results of tiny-scale VMamba (VMamba-T) for semantic segmentation on ADE20K. ‘SS’ and ‘MS’ denote single-scale and multi-scale testing, respectively. UperNet is utilized as the segmentation framework.}
\label{seg_result}
\captionsetup{skip=4pt}
\end{table}

\begin{table}[t]
\centering
\small
\renewcommand{\arraystretch}{0.9}
\begin{tabular}{c|c|cccc}
\toprule
Bit & Method & AP$^{\text{box}}$ & AP$^{\text{box}}_{\text{50}}$ & AP$^{\text{mask}}$ & AP$^{\text{mask}}_{\text{50}}$ \\ 
\midrule
\multirow{3}{*}{FP} & Swin-T & 42.7 & - & 39.3 & - \\ 
  & ConvNeXt-T & 44.2 & - & 40.1 & - \\ 
 & VMamba-T & 47.3 & 69.3 & 42.7 & 66.4 \\ 
\midrule
\multirow{4}{*}{2}& PQF    & 44.6 & 66.5 & 40.6 & 63.7\\ 
& DKM   & 45.7 & 68.2 & 41.5 & 64.8\\  
& VQ4DiT    & 45.5 & 67.9 & 41.4 & 64.7\\ 
& ViM-VQ (Ours)  & \textbf{46.0} & \textbf{68.8} & \textbf{41.8} & \textbf{65.7}\\
\midrule
\multirow{4}{*}{1}& PQF   & 33.1 & 53.0 & 31.2 & 50.5 \\ 
& DKM   & 37.5 & 58.0 & 35.4 & 55.4\\  
& VQ4DiT   & 36.6 & 56.9 & 34.0 & 54.3\\ 
& ViM-VQ (Ours)  & \textbf{40.9} & \textbf{60.7} &  \textbf{35.8} & \textbf{57.9} \\ 
\bottomrule
\end{tabular}
\caption{Quantization results of tiny-scale VMamba (VMamba-T) for object detection and instance segmentation on MSCOCO. AP$^{\text{box}}$ and AP$^{\text{mask}}$ denote box AP and mask AP, respectively. Mask R-CNN is utilized as the detection framework.}
\label{det_result}
\captionsetup{skip=4pt}
\vspace{-0.2cm}
\end{table}

\subsection{Semantic Segmentation}
As shown in Table~\ref{seg_result}, our method demonstrates strong performance on the ADE20K semantic segmentation benchmark. At 2-bit quantization, ViM-VQ achieves an mIoU of 44.4\% (single-scale) and 45.4\% (multi-scale), attaining performance comparable to full-precision Swin-T and ConvNeXt-T. Even under aggressive 1-bit quantization, ViM-VQ maintains an mIoU above 40\%, demonstrating its robustness in extremely low-bit settings.

\subsection{Object Detection and Instance Segmentation}
For object detection and instance segmentation on the MSCOCO dataset, ViM-VQ also achieves excellent results, as detailed in Table~\ref{det_result}. At 2-bit quantization, ViM-VQ consistently outperforms full-precision Swin-T and ConvNeXt-T, achieving 46.0\% box AP and 41.8\% mask AP. Even under the extreme condition of 1-bit quantization, ViM-VQ effectively mitigates performance degradation, still obtaining a high 40.9\% box AP and 35.8\% mask AP.

\begin{table}[t]
    \centering
    \small
    \renewcommand{\arraystretch}{0.9}
    \begin{tabular}{c|c|c}
    \toprule
    Bit & Method & Top-1 Acc (\%) \\
    \midrule
         FP &  Vim-T & 76.07  \\
          \midrule
          \multirow{3}{*}{2}
          &  Baseline ($\text{VQ} ^ \dagger $) & 33.18  \\
          &  + 1) Combination Optimization  & 71.42 (38.24$\uparrow$)  \\
          &  + 2) Incremental Quantization (Ours)  & \textbf{72.17} (0.75$\uparrow$)  \\
    \bottomrule
    \end{tabular}
    \caption{Ablation results for ViM-VQ for 2-bit quantization of tiny-scale Vision Mamba (Vim-T) on ImageNet-1K. $\text{VQ} ^ \dagger $ represents basic vector quantization.}
    \label{ablation}
    \captionsetup{skip=4pt}
\end{table}

\begin{table}[t]
    \centering
    \small
    \renewcommand{\arraystretch}{0.9}
    \begin{tabular}{c|c|ccccc}
    \toprule
     $\mathbf{C}$ & Method & $n$ & $E$ & Mem & Acc-C & Acc-I \\
    \midrule
    \multirow{5}{*}{\begin{tabular}[c]{@{}c@{}c@{}} 256 \\ $\times$ \\ 4 \end{tabular}}
     & DKM & 256 & 14 & 25 GB & 71.98 & 71.25 \\
     & VQ4DiT  & 4 & 2 & 11 GB & 71.33 & 71.04 \\
     & ViM-VQ  & \textbf{4} & \textbf{2} & \textbf{11 GB} & \textbf{72.17} & \textbf{72.17}  \\
     & ViM-VQ  & 16 & 4 & 12 GB & \textbf{72.31} & \textbf{72.31}  \\
     & ViM-VQ  & 64 & 8 & 14 GB & \textbf{72.34} & \textbf{72.34}  \\
    \bottomrule
    \end{tabular}
    \caption{Calibration efficiency of 2-bit quantization of tiny-scale Vision Mamba (Vim-T) on ImageNet-1K. $\mathbf{C}$, $n$, $E$, 'Mem' represent the codebook size, the number of candidate codewords, the final epoch, and the memory usage for calibration, respectively. 'Acc-C' and 'Acc-I' represent the Top-1 accuracy (\%) during calibration and inference.}
    \label{metric}
    \captionsetup{skip=4pt}
\end{table}

\begin{table}[t]
\small
\renewcommand{\arraystretch}{0.9}
\centering
\begin{tabular}{c|c|cccc}
    \toprule
    Bit & Method  & Resolution & Size (MB)  & CUDA  & Time \\
    \midrule
    FP & Vim-B  & 224$\times$224  & 1452 & n/a & 336s \\
    \midrule
    \multirow{2}{*}{\shortstack{
    2}}& \multirow{2}{*}{\shortstack{
    ViM-VQ}}     & 224$\times$224 & 92 & $\times$  & 357s \\
    &            & 224$\times$224 & 92 & $\checkmark$ & \textbf{328s} \\
    \bottomrule
\end{tabular}
\caption{Inference time of 2-bit base-scale Vision Mamba (Vim-B) on ImageNet-1K validation set on a single NVIDIA A6000 GPU. 'Size' represents the model size. 'CUDA' denotes whether the CUDA kernel for matrix–vector multiplication is used. }
\label{time}
\captionsetup{skip=4pt}
\vspace{-0.1cm}
\end{table}

\subsection{Ablation Study} 
Table~\ref{ablation} presents an ablation study on the contribution of each component within our proposed ViM-VQ framework. The basic VQ method suffers from substantial quantization errors due to weight outliers, leading to a significant accuracy drop to 33.18\%. By introducing the differentiable convex combination optimization algorithm, the accuracy improves by 38.24 percentage points, surpassing the performance of both DKM and VQ4DiT. This result highlights that updating both the convex combinations and the convex hulls effectively facilitates the search for the optimal codewords. Furthermore, incorporating the incremental vector quantization strategy provides an additional 0.75 percentage point improvement, demonstrating that the truncation error induced during the confirmation of optimal codewords is effectively mitigated.

As shown in Table~\ref{metric}, we report the calibration efficiency for 2-bit quantization of Vim-T on ImageNet-1K. DKM requires the most epochs and GPU memory during calibration, which can easily lead to out-of-memory issues. Although VQ4DiT significantly reduces computational overhead, its performance remains suboptimal due to the limited scope of candidate codewords. Moreover, DKM and VQ4DiT neglect the truncation error, resulting in a noticeable decrease in inference accuracy compared to calibration. In contrast, ViM-VQ outperforms DKM and VQ4DiT while maintaining performance consistency, indicating that it strikes a better balance between memory usage, calibration time, and final accuracy.

We develop a customized CUDA kernel for efficient matrix–vector multiplication. In our implementation, the codebooks are stored in constant GPU memory, while the assignments are fully loaded into CUDA blocks, enabling each thread to decode its corresponding assignment in parallel. The input is sequentially multiplied by the decoded weights, with the results accumulated in registers. As shown in Table~\ref{time}, our customized CUDA kernel reduces inference time from 357s to 328s, nearly matching the inference speed of the original, full-precision ViM.

\section{Conclusion}
\label{sec:conclusion}
In this work, we identify and address key challenges encountered when applying vector quantization (VQ) to Visual Mamba networks (ViMs). These challenges include severe quantization errors caused by significant weight outliers, as well as the high calibration costs, suboptimal performance, and truncation errors of existing VQ methods. To overcome these issues, we propose ViM-VQ, a post-training VQ method specifically tailored for ViMs. ViM-VQ incorporates a fast convex combination optimization algorithm for efficient optimal codeword search and an incremental vector quantization strategy to confirm optimal codewords while mitigating truncation errors. Experimental results demonstrate that ViM-VQ significantly outperforms existing approaches across diverse visual tasks, offering a practical solution for deploying ViMs on edge devices.

\section{Acknowledgements}
This work was supported by the National Natural Science Foundation of China (Grant No. 62274142), Sino-German Mobility Programme (Grant No. M-0499), and Zhejiang University - Vivo Information Technology Joint Research Center.

{
    \small
    \bibliographystyle{ieeenat_fullname}
    \bibliography{main}
}

\end{document}